\documentclass[fleqn,10pt]{wlscirep}
\usepackage[utf8]{inputenc}
\usepackage[T1]{fontenc}
\usepackage[numbers]{natbib}
\usepackage{times}
\usepackage{latexsym}
\usepackage{xspace}
\usepackage{hyperref}
\usepackage{wrapfig}
\usepackage{graphicx}
\usepackage{subcaption}

\usepackage[T1]{fontenc}

\usepackage[utf8]{inputenc}

\usepackage{microtype}

\usepackage{inconsolata}

\usepackage{graphicx}

\usepackage{multirow}
\usepackage{adjustbox}
\usepackage{makecell}
\usepackage{color}
\usepackage{soul}
\usepackage{bm}
\usepackage{enumitem}

\usepackage{tabularx}  
\usepackage{graphicx}
\usepackage{booktabs}
\usepackage[most]{tcolorbox}
\usepackage{mdframed} 
\usepackage{array} 
\usepackage{xcolor}  
\usepackage[left]{lineno}

\lstdefinestyle{datalogstyle}{
        basicstyle={\tt \scriptsize},  %
	xleftmargin={6pt},
        xrightmargin={6pt},
        columns=flexible,
        breakindent=0pt,
        breaklines=true, 
	frame=tb,
	stepnumber=1,
	firstnumber=1,
	numberfirstline=true,
	tabsize=2,
	extendedchars=true,
	breaklines=true,
	columns=fullflexible,
	keepspaces=true,
	escapeinside={@}{@},
	firstnumber=last,
	captionpos=b, 
	commentstyle=\color{black!65},
	numberstyle=\tiny\color{black!65},
	stringstyle=\color{codepurple},
	breakatwhitespace=false, 
	keepspaces=true,              
        mathescape=true, 
	numbersep=5pt,                  
	showspaces=false,                
	showstringspaces=false,
	showtabs=false,
	aboveskip={0.8\baselineskip},
	belowskip={0.2\baselineskip},
}

\usepackage{xcolor}
\usepackage{listings}
\usepackage{tcolorbox}

\definecolor{mygray}{RGB}{169,169,169}
\definecolor{myblue}{RGB}{0,102,204}
\definecolor{mygreen}{RGB}{0, 128, 0}
\definecolor{myred}{RGB}{255, 0, 0} 
\usepackage{colortbl}

\lstdefinestyle{customlatex}{
  basicstyle=\ttfamily\small,
  keywordstyle=\color{myblue},
  commentstyle=\color{mygray},
  stringstyle=\color{myred},
  breaklines=true,
  frame=single
}

\title{Are LLM-generated plain language summaries truly understandable? A large-scale crowdsourced evaluation}

\author[1]{Yue Guo, MBBS, PhD}
\author[2]{Jae Ho Sohn, MD, MS}
\author[3]{Gondy Leroy, PhD}
\author[4]{Trevor Cohen, MBChB, PhD, FACMI}
\affil[1]{School of Information Sciences, University of Illinois Urbana-Champaign, Champaign, IL}
\affil[2]{Department of Radiology and Biomedical Imaging, University of California, San Francisco, CA}
\affil[3]{Eller College of Management, University of Arizona, Tucson, AZ}
\affil[4]{Biomedical Informatics and Medical Education, University of Washington, Seattle, WA}

\affil[*]{Corresponding author: Yue Guo (yueg@illinois.edu)}

\begin{abstract}  
Plain language summaries (PLSs) are essential for facilitating effective communication between clinicians and patients by making complex medical information easier for laypeople to understand and act upon. Large language models (LLMs) have recently shown promise in automating PLS generation, but their effectiveness in supporting health information comprehension remains unclear. Prior evaluations have generally relied on automated scores that do not measure understandability directly, or subjective Likert-scale ratings from convenience samples with limited generalizability. To address these gaps, we conducted a large-scale crowdsourced evaluation of LLM-generated PLSs using Amazon Mechanical Turk with 150 participants. We assessed PLS quality through subjective Likert-scale ratings focusing on simplicity, informativeness, coherence, and faithfulness; and objective multiple-choice comprehension and recall measures of reader understanding. Additionally, we examined the alignment between 10 automated evaluation metrics and human judgments. 
Our findings indicate that 
while LLMs can generate PLSs that appear indistinguishable from human-written ones in subjective evaluations, human-written PLSs lead to significantly better comprehension.
Furthermore, automated evaluation metrics fail to reflect human judgment, calling into question their suitability for evaluating PLSs. 
This is the first study to systematically evaluate LLM-generated PLSs based on both reader preferences and comprehension outcomes. Our findings highlight the need for evaluation frameworks that move beyond surface-level quality and for generation methods that explicitly optimize for layperson comprehension.
\end{abstract}  

\begin{document}

\flushbottom
\maketitle

\section{Introduction}


Timely and accurate access to scientific study results is essential for informing the public, fostering trust, and ensuring transparency in the research process \cite{ravinetto2023responsible}. Recognizing this, researchers, and funding agencies such as the National Institutes of Health, emphasize the ethical responsibility of making study findings publicly accessible \cite{NIH2025}. One key approach is the use of plain language summaries (PLSs), which provide concise explanations of research findings designed for audiences without medical training. Many publishers now require PLSs when submitting research manuscripts, reinforcing their importance in science communication \cite{Guo2022RetrievalAO}.  

The development of automated methods for generating abstracts of scientific literature dates back to at least the late 1950s, when early research on abstractive summarization began \cite{luhn1958automatic}. However, the development of automated methods for generating PLSs - which by definition require translating from specialized content into lay language - emerged more recently, beginning with the use of the BART model \cite{lewis2019bart}, a sequence-to-sequence pre-trained transformer designed for natural language generation, translation, and comprehension. By fine-tuning BART on paired biomedical abstracts and human-authored PLSs \cite{guo2021automated, devaraj2021paragraph}, researchers demonstrated its ability to generate summaries of acceptable quality to human reviewers.   Recent advancements in large language models (LLMs), such as LLaMA \cite{touvron2023llama}  and GPT \cite{luo2023chatgpt}, have shown potential in automating PLS generation. These models can generate fluent and readable text, and prior studies indicate that LLM-generated PLSs generally achieve high simplicity scores in human evaluations \cite{Guo2022RetrievalAO, you2024uiuc_bionlp, goldsack2024overview, leroy2024text}. 

However, evaluating PLS quality is challenging as it requires balancing comprehensibility and accuracy. Unlike the traditional summarization task, which prioritizes brevity and faithfulness, PLS evaluation must also account for simplicity, informativeness, and coherence \cite{Guo2022RetrievalAO, cohen2025coherence}. Current assessment methods of PLS rely on automated metrics and human evaluation, both of which have significant limitations. Automated evaluation typically relies on summarization and simplification metrics \cite{jain2021summarization, ondov2022survey}, such as ROUGE \cite{lin2004rouge} and BLEU \cite{sulem2018bleu}. These metrics primarily measure lexical (n-gram) overlap between the generated text and reference text (e.g., human-authored PLS) rather than assessing the quality of the generated output. A higher n-gram overlap does not necessarily indicate a well-formed PLS, as these metrics fail to capture coherence, informativeness, and faithfulness \cite{Guo2023APPLSEE}. Furthermore, lower n-gram overlap does not necessarily indicate a poorly-formed summary, as there are many different ways to express the same ideas in plain language. 
Human evaluation, though considered the gold standard, also has limitations. Subjective Likert-scale ratings are influenced by cognitive biases, cultural backgrounds, and personal experiences \cite{hirschberg2003experiments, wiebe2004learning, haselton2015evolution}, leading to variability in interpretation. Additionally, prior studies have relied upon university-based samples of evaluators \cite{guo2021automated, Guo2022RetrievalAO}, which do not fully represent the intended PLS audience -- individuals without formal medical training -- limiting generalizability. 


To address these gaps, we conducted the first large-scale crowdsourced evaluation of LLM-generated PLSs using Amazon Mechanical Turk (MTurk), with 150 participants in total. We developed a comprehensive evaluation framework that integrates both \textit{subjective} and \textit{objective} measures. The former capture rater \textit{preferences}, while the latter measure rater \textit{comprehension} of the PLSs concerned. Within this framework, subjective evaluation is conducted through Likert-scale ratings assessing simplicity, informativeness, coherence, and faithfulness, while objective evaluation includes multiple-choice comprehension assessments and recall-based free-text responses. By incorporating diverse evaluation methods, this framework provides a more robust assessment of LLM-generated PLSs compared to prior work. By applying it, one can systematically examine whether LLMs can generate PLSs that adhere to specific stylistic and content preferences, and assess the extent to which these PLSs facilitate reader comprehension. 

Beyond human evaluation, we present an analysis of the applicability of automated evaluation metrics for PLS assessment. 
Previous research has tested automated scores using perturbation-based methods \cite{guo2022cells}. However, their alignment with human assessments remains unverified. Our study provides the first large-scale analysis of this alignment, offering insights into the strengths and limitations of existing automated evaluation metrics for PLSs.  

In summary, our contributions are as follows:  
\begin{itemize}  
    \item Constructing a large-scale, high-quality dataset of LLM-generated and human-authored PLSs paired with diverse human evaluations, including Likert ratings, multiple-choice questions, and recall responses.
    \item Investigating whether LLMs can generate understandable and high-quality PLSs that align with human preferences.  
    \item Assessing whether LLM-generated PLSs facilitate reader comprehension, as estimated using multiple choice questions and recall rates. 
    \item Examining the extent to which automated evaluation scores correspond with both human preferences and reader comprehension, providing insights into their reliability for PLS assessment.  
\end{itemize}

\section{Results}
To compare the differences between LLM- and human-generated PLSs, we use the CELLS dataset \cite{Guo2022RetrievalAO}, which contains over 63,000 pairs of human-authored scientific abstracts and their corresponding human-written PLSs. From this dataset, we randomly sampled 50 abstract-PLS pairs. For each sampled pair, we generated six LLM-based PLS variants from the scientific abstracts, each optimized for a distinct criteria: simplification, informativeness, coherence, faithfulness, all combined, and no specific optimization (see \S\ref{sec:method} for details). This design enables a controlled comparison between human-written summaries and their LLM-generated counterparts across different generation strategies. 

Each pair (scientific abstract and corresponding LLM- or human-generated PLS) was independently annotated by five crowd workers. A total of 1,750 annotations were initially collected. After excluding responses with missing values, we retained 1,346 complete annotations for the subsequent analysis. Considering previously documented concerns regarding data quality from MTurk \cite{beck2023quality}, we implemented additional quality controls. Specifically, annotations were excluded if participants answered corresponding attention question incorrectly, or if their completion time fell outside the 25\textsuperscript{th} to 75\textsuperscript{th} percentile range. After applying these quality control measures, 459 annotation pairs (34.1\%) remained and were utilized as a more rigorous evaluation subset.

We begin by reporting the characteristics of crowdsourced participants to provide context about the evaluation population (\S\ref{sec:human_character}). We then present linguistic and readability analyses (\S\ref{sec:readability_score}), followed by subjective evaluations (\S\ref{sec:subjective}) and objective comprehension-based assessments (\S\ref{sec:objective}) across LLM- and human-generated PLSs, using the full set of 1,346 annotations. These analyses address whether differences between PLS versions are detectable at the group level. 
Next, we examine the relationship between human quality ratings and comprehension performance (\S\ref{sec:sub_not_obj}), followed by an analysis of how well automated evaluation metrics predict human comprehension scores (\S\ref{sec:automated}) using the rigorous evaluation subset. Finally, we investigate the influence of individual-level human factors on comprehension outcomes (\S\ref{sec:human_factor}).

\subsection{Crowdsourced workers’ characteristics} \label{sec:human_character}

\begin{table}[h]
    \centering
    \small
    \caption{Demographic characteristics of crowdsourced workers (N = 150).}
    \label{tab:annotator_characteristics}
    \begin{minipage}{0.48\linewidth}
        \centering
        \caption*{General Characteristics}
        \begin{tabular}{@{}ll@{}}
            \toprule
            \textbf{Characteristic} & \textbf{Count (\%)} \\
            \midrule
            Age (Mean $\pm$ SD) & 37 $\pm$ 10 \\
            \midrule
            Education Level &  \\
            \hspace{5mm} No High School & 1 (1\%) \\
            \hspace{5mm} High School & 23 (15\%) \\
            \hspace{5mm} Associate’s Degree & 13 (9\%) \\
            \hspace{5mm} Bachelor’s Degree & 87 (58\%) \\
            \hspace{5mm} Master’s Degree & 21 (14\%) \\
            \hspace{5mm} PhD & 5 (3\%) \\
            \midrule
            English Spoken at Home &  \\
            \hspace{5mm} Always & 142 (95\%) \\
            \hspace{5mm} Sometimes & 6 (4\%) \\
            \hspace{5mm} Never & 2 (1\%) \\
            \bottomrule
        \end{tabular}
    \end{minipage}
    \hfill
    \begin{minipage}{0.48\linewidth}
        \centering
        \caption*{Social Characteristics}
        \begin{tabular}{@{}ll@{}}
            \toprule
            \textbf{Characteristic} & \textbf{Count (\%)} \\
            \midrule
            Ethnicity &  \\
            \hspace{5mm} Hispanic or Latino & 19 (13\%) \\
            \hspace{5mm} Not Hispanic or Latino & 122 (81\%) \\
            \hspace{5mm} Unknown & 7 (5\%) \\
            \midrule
            Gender &  \\
            \hspace{5mm} Female & 72 (48\%) \\
            \hspace{5mm} Male & 66 (44\%) \\
            \hspace{5mm} Non-binary & 4 (3\%) \\
            \hspace{5mm} Gender Non-conforming & 1 (1\%) \\
            \hspace{5mm} Transgender Female & 5 (3\%) \\
            \hspace{5mm} Transgender Male & 1 (1\%) \\
            \hspace{5mm} Prefer not to answer & -- \\
            \midrule
            Race &  \\
            \hspace{5mm} American Indian / Alaskan Native & 2 (1\%) \\
            \hspace{5mm} Asian & 10 (7\%) \\
            \hspace{5mm} Black / African American & 11 (7\%) \\
            \hspace{5mm} Hawaiian / Pacific Islander & -- \\
            \hspace{5mm} White & 124 (83\%) \\
            \hspace{5mm} Unknown & 3 (2\%) \\
            \bottomrule
        \end{tabular}
    \end{minipage}
\end{table}

We recruited participants through MTurk with the inclusion criteria that they must be (1) proficient in English and (2) reside in the United States. To ensure lay reader perspectives, we excluded individuals who (1) had participated in medical training or shadowing in a hospital, or (2) had completed advanced (graduate level) biology courses.

A total of 500 qualified annotators were identified, each randomly assigned one of seven evaluation batches. Annotators could complete up to 20 pairs, with compensation set at \$1 per pair. In total, 150 annotators participated in the study, providing responses to our tasks. The demographic characteristics of the annotators are summarized in Table~\ref{tab:annotator_characteristics}. The average participant age at recruitment is 37, and 58\% hold a bachelor’s degree. 95\% percent speak English at home, while 48\% identify as female. Regarding ethnicity and race, 81\% of participants are not Hispanic or Latino, and 83\% identify as White, with additional representation from other racial backgrounds. Overall, the study sample reflects a diverse participant pool. The study was considered exempt by the 
our Institutional Review Board (IRB). Electronic informed consent was obtained from each study participant.

\subsection{LLM- and human-generated PLSs' characteristics} \label{sec:readability_score}

\begin{figure}
    \centering
    \includegraphics[width=1\textwidth]{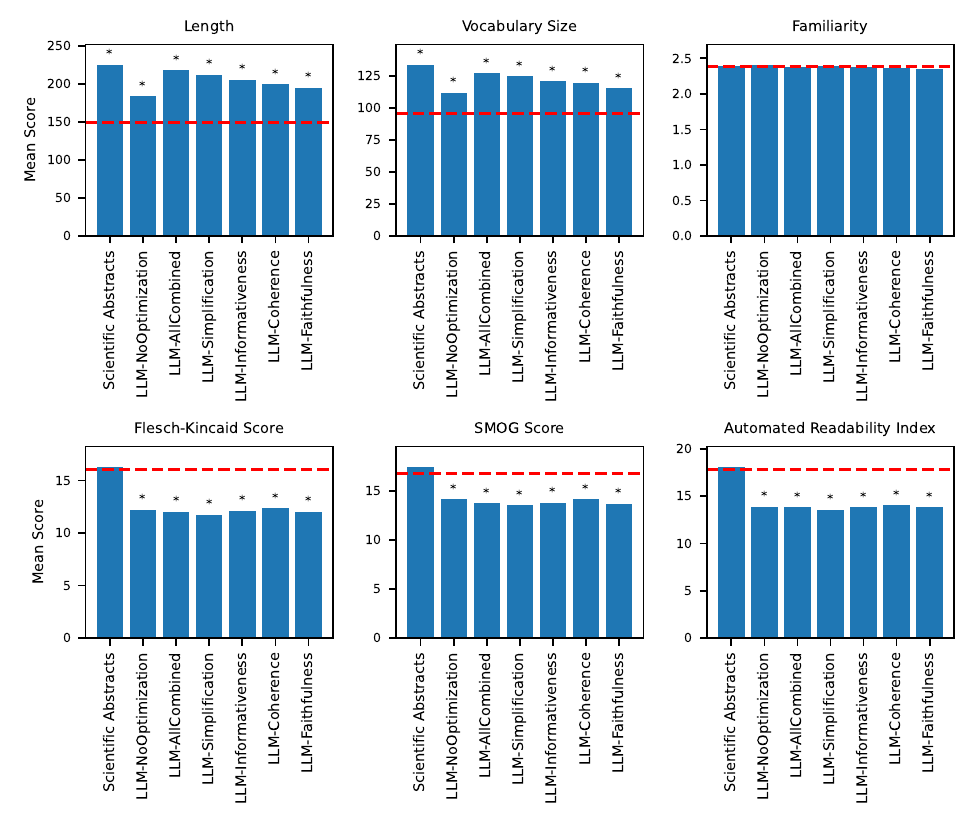}
    \caption{Comparison of linguistic and readability metrics across human-written and LLM-generated PLS under various optimization strategies (N = 1346). Metrics include paragraph length, vocabulary size, word familiarity, and three standard readability scores (Flesch-Kincaid, SMOG, Automated Readability Index). The \textcolor{red}{red} dashed line indicates the mean score for human-written PLS. Asterisks (*) denote statistically significant differences (paired t-test, p < 0.05) between each system and the human-written PLS. $p$-values are adjusted for multiple comparisons using the Benjamini--Hochberg correction. }
    \label{fig:all_variables_bh}
\end{figure}

To compare LLM-generated PLSs with human-written ones, we took the human-authored PLSs as the baseline and evaluated the outputs from various LLM optimization strategies against them. For broader context, we also included the aligned human-written scientific abstracts in the analysis.

As shown in Figure~\ref{fig:all_variables_bh}, in terms of paragraph length, scientific abstracts are the longest, while human-written PLSs (indicated by the red dashed line) are the shortest. LLM-generated PLSs fall in between. We also examined vocabulary size, measured as the number of unique words in each text—a standard linguistic feature for assessing lexical richness. As expected, scientific abstracts have the largest vocabulary size, followed by LLM-generated PLSs, with human-written PLSs being the most limited. This pattern is expected, as vocabulary size is highly correlated with overall paragraph length. To assess lexical familiarity, we calculated the percentage of words that fall within the 1,000 most common English words, a commonly used proxy for text familiarity \cite{leroy2018next}. There is no significant difference in this measure between human-written and LLM-generated PLSs.

We further computed three widely used readability scores: Flesch-Kincaid Grade Level, SMOG Index, and the Automated Readability Index, where lower values indicate higher readability. As expected, human-written scientific abstracts received higher (i.e., less readable) scores than their corresponding PLSs; however, the difference was not statistically significant. This suggests that readability metrics may not effectively distinguish between scientific and lay language, or that human-authored PLSs are not inherently more readable. In contrast, all LLM-generated PLSs exhibited significantly lower readability scores than human-written PLSs, indicating higher estimated readability. This pattern suggests that LLMs may implicitly optimize for surface-level readability features during generation, even without explicit prompts to do so.

\subsection{LLM- and human-generated PLSs have similar subjective ratings} \label{sec:subjective}  

\begin{figure}
    \centering
    \includegraphics[width=0.8\textwidth]{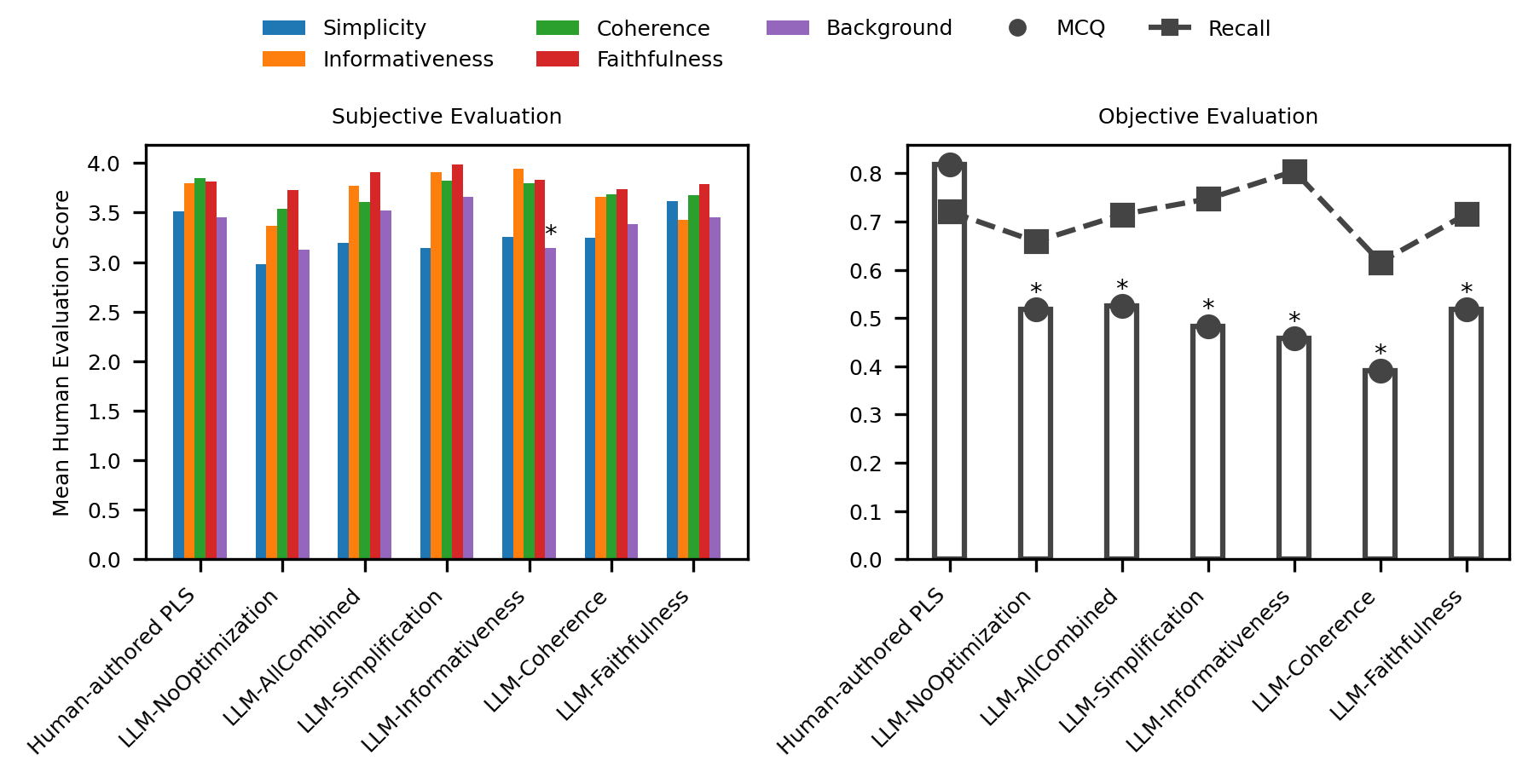}
    \caption{Subjective and objective evaluation of human-written and LLM-generated PLS. The left panel (''Subjective Evaluation'') shows mean human Likert-scale ratings across five quality dimensions: simplicity, informativeness, coherence, faithfulness, and providing necessary background knowledge (N = 1346). The right panel (''Objective Evaluation'') presents performance on comprehension tasks: multiple-choice questions (MCQ) accuracy and recall rate. Colored bars represent subjective dimensions; hatched bars indicate MCQ and Recall scores, with distinct patterns. Asterisks (*) mark statistically significant differences to the human-written PLS baseline (paired t-test, p < 0.05). $p$-values are adjusted for multiple comparisons using the Benjamini--Hochberg correction. }
    \label{fig:llm_human_all_bh}
\end{figure}

For the Likert-scale evaluations of simplicity, informativeness, coherence, factuality, and the inclusion of additional background information, we used human-authored PLS as the baseline. As illustrated in Figure~\ref{fig:llm_human_all_bh} left panel, none of the LLM-generated PLSs were rated significantly higher than the human-written versions on any dimension. PLSs optimized for informativeness were rated significantly lower than human-written PLSs on inclusion of necessary background information (i.e. whether the necessary and relevant background information added to improve your understanding). For the remaining dimensions, no statistically significant differences were observed between the LLM-generated and human-authored PLSs. These results suggest that, from a subjective standpoint, participants generally perceived LLM-generated content as comparable to human-written summaries. However, the lack of consistent differences across the six LLM optimization strategies indicates that LLMs did not reliably adhere to prompts designed to emphasize specific evaluation criteria. Notably, PLSs explicitly instructed to prioritize simplicity received the \textit{lowest} simplicity scores, further underscoring this limitation.

\subsection{LLM-generated PLSs are harder to understand}  \label{sec:objective}  
However, as shown in Figure~\ref{fig:llm_human_all_bh} right panel, objective comprehension measures tell a different story. For the multiple-choice questions, which remained identical across all PLS versions, participants exhibited significantly higher accuracy when answering questions based on human-authored PLS compared to LLM-generated versions. This finding suggests that, while subjective evaluations show little differentiation, LLM-generated PLSs are less effective in supporting comprehension. While participants showed slightly higher recall scores with some LLM-generated PLSs, the difference was not statistically significant. The discrepancy between subjective perception and multiple-choice questions in particular underscores the importance of incorporating such evaluations as a quantitative measure of comprehensibility, rather than relying solely on subjective assessments. Future research should explore refining LLM-generated PLS to improve their ability to convey information effectively while maintaining clarity and coherence.

\begin{table}[ht]
\centering
\caption{Mixed-effects model results for human ratings predicting normalized multiple-choice question comprehension scores in rigorous evaluation subset (N = 459). $p$-values are adjusted for multiple comparisons using the Benjamini--Hochberg correction. Metrics with $p$-values less than 0.05 are considered statistically significant.}
\begin{tabular}{lccccc}
\toprule
\textbf{Metric} & \textbf{Coef.} & \textbf{$p$-value} & \textbf{95\% CI Lower} & \textbf{95\% CI Upper} & \textbf{Significant} \\
\midrule
Faithfulness     & 0.048 & 0.014 & 0.016 & 0.079 & Yes \\
Background       & 0.045 & 0.016 & 0.013 & 0.078 & Yes \\
Coherence        & 0.040 & 0.019 & 0.009 & 0.071 & Yes \\
Informativeness  & 0.039 & 0.020 & 0.007 & 0.072 & Yes \\
Simplicity       & 0.037 & 0.029 & 0.004 & 0.071 & Yes \\
\bottomrule
\end{tabular}
\label{rating_mcq_mixed_small_bh}
\end{table}

\subsection{Subjective ratings predict comprehension}  \label{sec:sub_not_obj}  
We examined the association between human subjective ratings and comprehension performance using linear mixed-effects models with multiple-choice questions accuracy as the outcome. Each model included a single standardized rating variable (e.g., faithfulness, simplicity) as a fixed effect, and random intercepts for both participants and abstracts to account for individual- and item-level variation.
In the human evaluation results shown in Table~\ref{rating_mcq_mixed_small_bh}, all five subjective quality ratings -- faithfulness, simplicity, coherence, informativeness, and background -- were significantly associated with comprehension performance (\( p < .05 \), after multiple comparison correction). Among them, faithfulness showed the strongest predictive relationship (\( \beta = 0.048 \)), followed closely by background and coherence (all \( \beta \geq 0.04 \)), suggesting that summaries perceived as factually accurate, contextually informative, and logically structured tend to facilitate better comprehension. These findings underscore the complementary roles of different subjective dimensions in shaping reader understanding and suggest that, when evaluation resources are limited, prioritizing faithfulness, background information, and coherence may offer the greatest return for supporting comprehension in PLSs. Results from the full dataset, presented in Appendix Table~\ref{rating_mcq_mixed_whole_bh}, are consistent with those from the rigorous evaluation subset.

Notably, the \textit{Background} rating asked participants whether necessary and relevant background information was added to the summary to improve their understanding. This further underscores the importance of elaborative explanation in PLSs~\cite{Guo2022RetrievalAO, srikanth2020elaborative}, which aims to enhance understanding by including information not present in the original scientific abstract, such as definitions, examples, or explanatory context.


\begin{table}[ht]
\centering
\caption{Mixed-effects model results for automated metrics predicting normalized multiple-choice question comprehension scores in rigorous evaluation subset (N=459). $p$-values are adjusted for multiple comparisons using the Benjamini--Hochberg correction. Metrics with $p$-values less than 0.05 are considered statistically significant.}
\begin{tabular}{lccccc}
\toprule
\textbf{Metric} & \textbf{Coef.} & \textbf{$p$-value} & \textbf{95\% CI Lower} & \textbf{95\% CI Upper} & \textbf{Significant} \\
\midrule
QAEval         & 0.039 & 0.035 & 0.013 & 0.065 & Yes \\
Coherence-bag  & 0.033 & 0.053 & 0.006 & 0.059 & No \\
Coherence-chn  & 0.032 & 0.053 & 0.006 & 0.058 & No \\
GPT-PPL        & 0.014 & 0.566 & -0.012 & 0.041 & No \\
BERTScore      & 0.011 & 0.655 & -0.015 & 0.037 & No \\
BLEU           & 0.007 & 0.717 & -0.019 & 0.034 & No \\
LENS           & 0.006 & 0.717 & -0.020 & 0.033 & No \\
ROUGE          & -0.001 & 0.961 & -0.027 & 0.025 & No \\
METEOR         & -0.010 & 0.655 & -0.036 & 0.016 & No \\
SARI           & -0.016 & 0.566 & -0.042 & 0.010 & No \\
\bottomrule
\end{tabular}
\label{auto_mcq_mixed_small_bh}
\end{table}

\subsection{Automated scores align poorly with human judgment} \label{sec:automated}   
Although human evaluation is the gold standard for assessing PLS quality, automated metrics are more scalable, cost-effective, and readily applicable during model development. To assess their utility, we examined the alignment between a range of widely used automated metrics in PLS evaluation. These include overlap-based metrics (ROUGE, BLEU, METEOR, SARI), model-based metrics (GPT-PPL, BERTScore, LENS), a question-answering (QA)-based metric (QAEval), and LLM-derived text coherence scores (Coherence-bag, and Coherence-chn). Implementation details are provided in \S\ref{sec:method}.

We also assessed the extent to which automated evaluation metrics predict comprehension performance using linear mixed-effects models, with normalized multiple-choice questions accuracy as the outcome. Each model included a single standardized automated metric as a fixed effect, with random intercepts for participants and abstracts to account for individual- and content-level variation. 
As shown in Table~\ref{auto_mcq_mixed_small_bh}, only one metric -- QAEval -- was significantly associated with comprehension performance after adjusting for multiple comparisons (\( \beta = 0.039 \), \( p = 0.035 \)). Coherence-based metrics (Coherence-bag and Coherence-chn), which capture paragraph-level and discourse-level structure respectively, showed comparable effect sizes but did not reach significance (\( p = 0.053 \)). Coherence-bag estimates the topical fit of each sentence relative to a summary-derived topic, while Coherence-chn additionally conditions on preceding sentences to model discourse flow. Traditional surface-form metrics such as BLEU, ROUGE, METEOR, GPT-PPL, and LENS were not significantly predictive of comprehension. These findings suggest that metrics explicitly targeting factual consistency, such as QAEval, may better reflect human understanding, while lexical overlap and perplexity-based metrics fail to capture the quality dimensions most relevant for supporting comprehension. Results from the full dataset, presented in Appendix Table~\ref{auto_mcq_mixed_whole_bh}, were directionally consistent with those from the rigorous evaluation subset.

\begin{figure}
    \centering
    \includegraphics[width=0.9\textwidth]{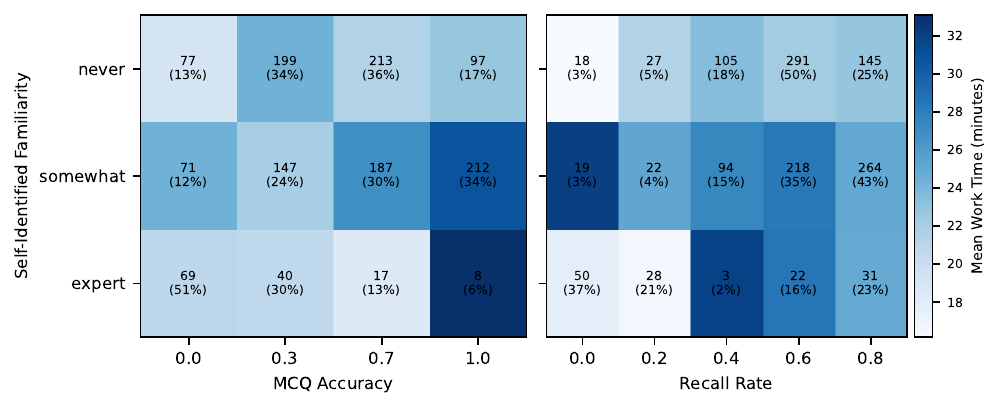}
\caption{Heatmaps of mean task time grouped by multiple-choice question (MCQ) accuracy (left) and recall rate (right), with rows representing self-identified familiarity (N = 1346). Color indicates average work time (in minutes). Each cell shows both the count of participants and the percentage within the corresponding familiarity group.}
    \label{fig:worktime}
\end{figure}

\subsection{Human factors in PLS evaluation} \label{sec:human_factor}
 
To better understand participants' behavior in evaluating PLS, we further analyzed their engagement patterns. As shown in Figure~\ref{fig:worktime}, in general, participants who spent more time on the task (darker blue) tended to achieve higher multiple-choice question accuracy and recall rates. Participants who reported no familiarity with the topic performed surprisingly well on multiple-choice questions, with 53\% answering more than two out of three correctly. This group also had a relatively even distribution of recall scores, with many achieving moderate recall performance. In contrast, the ''somewhat familiar'' group demonstrated the strongest overall comprehension: 64\% answered at least two multiple-choice questions correctly, and the majority achieved mid-to-high recall rates. This group represents the most balanced performance across both objective measures.

Interestingly, self-identified experts had the lowest multiple-choice question accuracy overall, with only 21\% answered more than two questions correctly, and their recall performance was more polarized, with a higher proportion falling in the low and high extremes. Despite their confidence, this group did not consistently outperform others. These results suggest that moderate familiarity may lead to greater engagement and comprehension, while overconfidence among self-identified experts may reduce performance. The data highlights the need to consider audience background when designing and evaluating plain language summaries.




\section{Discussion \& Conclusion}

This study had two primary goals: first, to evaluate whether LLMs can generate PLSs comparable to human-authored versions, and second, to develop a robust evaluation framework for assessing the quality of PLSs.

Our results demonstrate a clear gap between perceived and actual comprehension of LLM-generated PLSs. While participants rated these summaries similarly to human-written ones in terms of simplicity and coherence, objective measures such as multiple-choice question accuracy and free-text recall rate indicated significantly better understanding when participants read human-written PLSs. These findings suggest that subjective ratings may primarily reflect surface-level fluency rather than the effective communication of scientific content. This discrepancy aligns with recent studies in AI-generated persuasion, which show that personalized messages generated by LLMs are often perceived as more persuasive than traditional messaging in various domains \cite{matz2024potential}. Similarly, GPT-3-generated public health content was rated more persuasive than official CDC messaging \cite{karinshak2023working}. However, these assessments are based on subjective perception rather than behavioral outcomes. Our findings reinforce the importance of distinguishing between perceived quality and actual comprehension, especially in health communication contexts.

The study also sheds light on how PLS quality has been evaluated historically and the limitations of current practices. Prior to the widespread use of LLMs in PLS generation, most studies focused on user experience, specifically how accessible or understandable PLSs were perceived to be. These evaluations typically emphasized readability \cite{rakedzon2017automatic, martinez2020can, brehaut2011presenting}, user enjoyment or preference \cite{buljan2020comparison, buljan2020framing}, satisfaction with text length \cite{martinez2020can}, and usability judgments \cite{santesso2015summary, martinez2020can, brehaut2011presenting}. While some work incorporated comprehension measures such as multiple-choice or open-ended questions \cite{bredbenner2019video, buljan2018no}, these were less common.
With LLM-generated PLSs, evaluation has leaned heavily on readability scores and automated simplification metrics, often with small-scale human studies \cite{devaraj2021paragraph, guo2021automated}. Our findings highlight that these approaches are insufficient for capturing whether readers actually understand the content. To address this, we advocate for comprehension-centered evaluation protocols which including structured question answering as a standard component of human assessment.


Our analysis underscores the limitations of existing automated metrics in capturing comprehension-relevant aspects of summary quality. While traditional surface-form metrics such as BLEU, ROUGE, and METEOR are commonly used in PLS evaluation, our results show that they are not significantly predictive of reader comprehension. In contrast, QAEval demonstrated the strongest association with multiple-choice question accuracy, suggesting that QA-based evaluation methods are better aligned with human understanding, likely due to their sensitivity to factual correctness and content relevance. Coherence-based metrics, particularly Coherence-bag and Coherence-chn, also showed associations with comprehension outcomes, reinforcing the importance of paragraph-level and discourse-level coherence in supporting reader understanding. These findings highlight the potential of metrics grounded in factual consistency and discourse structure as more meaningful proxies for evaluating PLSs. They also point to a clear need for developing domain-adapted, comprehension-sensitive evaluation tools that move beyond surface-level text features and reflect the intended use of PLSs as bridges between scientific content and public understanding.

Participant characteristics also influenced comprehension outcomes. In our rigorous evaluation subset, which comprised 34\% of the total annotations, the overall result patterns were consistent with the full dataset, although the agreement between human and automated metrics was stronger. These findings suggest that incorporating attention checks and filtering responses based on completion time can improve data quality. In addition, most prior studies relied on small samples composed primarily of highly educated participants or domain experts \cite{martinez2020can, guo2021automated, kirkpatrick2017understanding}, which limits the generalizability of their findings. In contrast, our use of a more diverse participant pool from MTurk offers broader insights into how lay audiences interact with PLSs. Interestingly, participants with moderate familiarity with the topic outperformed self-identified experts, who tended to spend less time on the task and demonstrated lower comprehension scores. This highlights the importance of engagement and cognitive effort in understanding PLS content and underscores the need to account for user background and interaction behavior in future evaluations.

It is noteworthy that among all subjective quality dimensions, \textit{background} showed the second strongest association with comprehension performance in our mixed-effects models. This study represents the first large-scale crowdsourced evaluation to empirically demonstrate the benefits of including background information in PLSs. While prior qualitative studies have observed that PLS authors frequently introduce definitions, examples, or contextual explanations not found in the original scientific abstracts \cite{guo2021automated}, these transformations were often noted descriptively rather than evaluated systematically. Such additions are typically unnecessary for domain experts (the intended audience of scientific abstracts), but are essential for lay readers who may lack prior exposure to core concepts or terminology. Our findings provide quantitative evidence that including necessary and relevant background content significantly supports lay audience comprehension. This highlights the importance of designing PLSs not only for linguistic simplicity but also for conceptual accessibility, through deliberate inclusion of external contextual information.

This study makes several important contributions. It is the first to comprehensively compare six optimizations of LLM-generated PLSs to human-written versions in terms of knowledge gain. Our large-scale evaluation, which included 150 crowd-sourced participants and 1,346 annotated PLS-abstract pairs, provides a broader and more representative assessment than prior work. We also demonstrate the value of using a rigorous annotation subset and offer guidance on expected retention rates for future MTurk-based studies. Importantly, we present a framework that integrates both subjective and objective evaluations, paving the way for more meaningful benchmarks in PLS generation. These insights could inform the development of reinforcement learning objectives that prioritize comprehension over readability alone.

This study has several limitations. First, although we sampled 50 unique scientific abstracts from a large corpus, the limited number constrains the topical diversity of the evaluation. However, this sample size still exceeds that of prior PLS studies involving human evaluation \cite{guo2021automated, Guo2022RetrievalAO}. Second, while our use of MTurk provided a more demographically diverse participant pool than typical university-based convenience samples, it may still fall short of fully representing the general population in terms of literacy, health knowledge, or digital access. Finally, although human-written PLSs resulted in better performance on comprehension tasks, the underlying textual or structural factors contributing to this advantage remain unclear. Future work should explore which specific characteristics, such as sentence structure, lexical choice, or information ordering, are most effective in supporting layperson understanding.

Taken together, our results underscore the need to move beyond stylistic fluency in both the generation and evaluation of LLM-generated health communications. More robust evaluation frameworks grounded in comprehension and factual accuracy are essential to ensure that PLSs truly fulfill their purpose of informing and empowering non-expert audiences. Future research should explore generation strategies that directly optimize for human understanding and develop evaluation metrics capable of reliably capturing comprehension outcomes in real-world settings.

\section{Method} \label{sec:method}

\subsection{Dataset}
To ensure the quality of scientific abstracts and PLSs, we randomly sampled 50 abstract-PLS pairs from the CELLS dataset \cite{Guo2022RetrievalAO}, a large set of scientific abstracts paired with human-authored PLSs drawn from a range of biomedical journals. Following prior work using perturbation-based evaluation methods \cite{Guo2023APPLSEE}, we developed our subjective evaluation based on four key criteria: simplicity, informativeness, coherence, and faithfulness. To assess the ability of LLMs to optimize for these criteria, we also designed four targeted prompts, each aimed at enhancing a specific aspect of PLS quality. We used \texttt{GPT-4} for PLS generation \cite{OpenAI2024GPT4}, setting the maximum output length to 350 tokens, matching the maximum length of the 50 human-authored PLSs. All other parameters are set to their default values. Consequently, we generated and evaluated seven versions of PLSs: (1) human-authored, (2) criteria-agnostic, and oriented toward (3) informativeness, (4)simplification, (5) coherence, (6) faithfulness,and (7) all of these criteria.

The generation criteria used for \texttt{GPT-4} are defined as follows:  
\begin{itemize}
    \item \textbf{Informativeness}: The summary should comprehensively capture the essential elements of the original scientific text, such as the methodologies employed, primary findings, and conclusions drawn. It should effectively convey the core message without omitting critical details or introducing information not pertinent to the source material (i.e., hallucinations), as both can hinder reader comprehension.
    \item \textbf{Simplification}: The summary should present information in a manner that is easily interpretable and understandable to non-expert audiences. Prioritize the use of simple vocabulary and concise sentences while minimizing the use of jargon and technical terminology that may be unfamiliar to a lay audience. Add necessary and relevant background information if that would benefit the reader\'s understanding.
    \item \textbf{Coherence}: The summary should be logically structured, ensuring a clear and steady progression of ideas. Information should be presented in a well-ordered manner that promotes easy comprehension for the reader. Adhering to the original sentence order is recommended to maintain optimal coherence.
    \item \textbf{Faithfulness}: The summary should accurately align with the factual content of the source scientific text, including its findings, methods, and claims. Avoid substituting information or introducing errors, misconceptions, or inaccuracies that could mislead the reader or misrepresent the original author\'s intent.
\end{itemize}

\begin{lstlisting}[style=customlatex, caption={Prompt for \texttt{GPT-4} to generate PLS versions. Each input is a scientific abstract from the same set of 50 pairs.}]
@\textcolor{mygray}{System}@
Imagine you are a skilled scientific writer tasked with summarizing a scientific abstract in plain language. To achieve this, please adhere to the following steps: 
1. Read the provided scientific abstract carefully, paying close attention to the information it contains. 
2. Rewrite the scientific abstract as a plain language summary, ensuring that it aligns with the following criteria: @\textcolor{myred}{<criteria>}@.

@\textcolor{mygray}{User}@
Scientific abstract: @\textcolor{myblue}{<input>}@
Plain language summary: @\textcolor{mygreen}{<>}@
\end{lstlisting}

\subsection{Evaluation Procedure and Question Design}  

For the human evaluation, we prepared data in pairs consisting of an \textit{ABSTRACT} and a \textit{SUMMARY}. The ABSTRACT was the original scientific abstract, while the SUMMARY is a PLS, which could be one of seven versions, either human-authored or LLM-generated. To assess the quality and comprehensibility of the PLSs, we collected detailed participant responses using both subjective and objective measures.  

The LLM-PLSs were generated using \texttt{GPT-4}. To prevent GPT-4 from being biased toward its own generated text \cite{panickssery2024llm}, we used \texttt{LLaMA-3} to generate both attention-check and multiple-choice comprehension questions. Specifically, we employed the \texttt{Llama-3-70B} model. The model was quantized to 4 bits using OPTQ, as implemented in \url{https://github.com/PanQiWei/AutoGPTQ}. The maximum output length was set to 500 tokens. All other parameters were left at their default values. All generated multiple-choice and attention-check questions were manually reviewed to ensure quality and clarity.
The attention questions were created using the following prompt:  

\begin{quote}
\texttt{Create a multiple-choice question in plain language that examines the main topic of the text. The correct answer should be the most common noun in the text, with incorrect options consisting of similar-sounding but unrelated nouns. No explanation is needed; start with the question.}  
\end{quote}  

The comprehension questions were generated using the following prompt:  

\begin{quote}
\texttt{Create three multiple-choice questions in plain language that assess (1) the motivation of the study, (2) the methods used, and (3) the main results.}  
\end{quote}  

For the evaluation task, participants first reviewed a SUMMARY, and then completed the following tasks:  

\begin{itemize}  
    \item \textbf{Topic familiarity}: How familiar are you with the topic discussed in the SUMMARY? (Options: never encountered this topic, somewhat familiar, expert)  
    \item \textbf{Comprehension questions}: Three multiple-choice questions, generated by \texttt{LLaMA-3}, assessing participants' understanding of the study's motivation, methods, and results.  
    \item \textbf{Simplicity}: On a scale of 1 to 5, how easy was it to understand the SUMMARY?  
    \item \textbf{Coherence}: On a scale of 1 to 5, how well-structured and logically organized was the SUMMARY?  
    \item \textbf{Attention-check question}: To ensure data reliability, participants answer a multiple-choice question generated by LLaMA-3 that tests the main topic of the text.  
\end{itemize}  

After completing these initial tasks, the SUMMARY was removed from view. Participants then performed a recall task in which they wrote down as much information as they remember from the PLS, without copying it. In the next phase, participants were presented with both the SUMMARY and the original ABSTRACT and were asked to compare them across three dimensions:  

\begin{itemize}  
    \item \textbf{Informativeness}: On a scale of 1 to 5, how well did the PLS capture the content of the SOURCE?
    \item \textbf{Background information}: On a scale of 1 to 5, was necessary and relevant background information added in the SUMMARY to improve your understanding?  
    \item \textbf{Faithfulness}: On a scale of 1 to 5, how accurately did the SUMMARY align with the factual content of the SOURCE?
\end{itemize}  

We collected responses across multiple versions of the PLS, with each participant evaluating only one version. Participants were not informed as to the source or orientation of the summaries they reviewed.  This structured evaluation approach allowed us to assess the quality of PLSs comperehensively using both subjective and objective measures, for a robust analysis of their comprehensibility, accuracy, and informativeness.

\subsection{Automated Metrics}
Automated metrics for text simplification and summarization have been widely used in previous research to assess the quality of PLS.
While prior work \cite{Guo2023APPLSEE} has demonstrated that existing metrics have limitations in evaluating PLS quality using a perturbation-based testbed, no previous study has systematically examined the relationship between automated scores and human evaluation. To address this gap, we incorporate the same set of metrics used in our prior work\cite{Guo2023APPLSEE} to assess their alignment with human judgment of PLS based on four key criteria. The metrics are as follows:

\textbf{Overlap-based metrics} rely on $n$-gram overlap and are widely used due to their computational efficiency.

\begin{itemize}
    \item \textbf{ROUGE}\cite{lin2004rouge} evaluates textual similarity based on $n$-gram recall between generated summaries and reference texts. We report the average of ROUGE-1, ROUGE-2, and ROUGE-L.
    \item \textbf{BLEU} \cite{papineni2002bleu} measures $n$-gram precision of the generated text against reference summaries, incorporating a brevity penalty to discourage overly short outputs.
    \item \textbf{METEOR}\cite{banerjee2005meteor} improves upon BLEU by incorporating stemming, synonym matching, and an F-measure-based scoring approach that better accounts for recall.
    \item \textbf{SARI}\cite{xu2016optimizing} is designed specifically for text simplification tasks. It evaluates the quality of a simplified text by weighting the $n$-grams that are added, deleted, and retained relative to the source and target texts.
\end{itemize}

\textbf{Model-based metrics} leverage pretrained neural models to assess text qualities beyond lexical overlap.

\begin{itemize}
    \item \textbf{GPT-PPL} is typically computed using \texttt{GPT-2} and reflects fluency and coherence by calculating the exponentiated average negative log-likelihood of the tokens in a text. Lower perplexity scores indicate higher fluency and coherence. 
    \item \textbf{BERTScore}\cite{zhang2019bertscore} evaluates semantic similarity by computing F1 scores between contextualized embeddings of the hypothesis and reference texts, capturing meaning beyond lexical overlap.
    \item \textbf{LENS} \cite{maddela2022lens} is a reference-free metric for text simplification trained on human preference data. It uses an adaptive ranking loss during training to learn to score outputs based on edit operations such as splitting, paraphrasing, and deletion, favoring simplifications that better align with human judgments.
\end{itemize}

\textbf{QA-based metrics} evaluate content quality using a question-answering framework.

\begin{itemize}
    \item \textbf{QAEval} \cite{deutsch2021towards} assesses factual consistency by generating question-answer pairs from the reference text and testing whether a QA model can extract the correct answers from the generated summary. We report QAEval LERC scores, which measure semantic alignment between predicted and reference answers.
\end{itemize}

\textbf{LLM-derived text coherence scores} measure how semantically or probabilistically coherent a text is based on representations from LLMs. We evaluated two classes of coherence metrics derived from LLMs which were most strongly associated with accuracy of answers to multiple choice questions in \cite{cohen2025coherence}:  

\begin{itemize}
    \item \textbf{Bag/topical coherence} estimates the likelihood of each sentence given the overall topic of the passage. The topic is defined using a one-sentence summary generated by a BART-based summarization model \footnote{facebook/bart-large-xsum}. The log-likelihoods of sentences which computed using LLMs reflect their topical fit.
    \item \textbf{Chain/contextual coherence} extends the bag model by conditioning sentence probabilities not only on the topic but also on all preceding sentences in the text. This method captures discourse-level coherence by modeling the likelihood that a sentence logically follows from prior context using LLM-derived language probabilities.
\end{itemize}

To compute the automated evaluation scores, we used \texttt{Summeval} \cite{fabbri2021summeval} for ROUGE, BLEU, METEOR, and BERTScore (BERTScore hash: \texttt{bert-base-uncased\_L8\_no-idf\_version = 0.3.12}, Hugging Face Transformers v4.27.3). SARI scores were computed using \texttt{EASSE} \cite{alva-manchego-etal-2019-easse}. For GPT-PPL, we followed the implementation details provided at \href{https://huggingface.co/transformers/v3.2.0/perplexity.html}{this link}.

\subsection{Analysis}
In our analysis, we computed the average accuracy across the three multiple-choice questions. For the recall task, we measured token-level overlap between participants' written answers, capturing information recalled from the PLSs without direct copying, and the original PLS. Responses to Likert-scale questions were recorded numerically on a scale from 1 to 5. Paired t-tests were used to calculate the significance of observed differences, and the Benjamini--Hochberg Correction was performed to correct for multiple comparison.

To examine the relationship between human ratings, automated evaluation metrics, and comprehension performance, we fit a series of linear mixed-effects models using normalized multiple-choice question  accuracy as the dependent variable. Each model included a single standardized predictor (e.g., ROUGE, BERTScore, faithfulness) as a fixed effect. Because automated metrics are computed once per abstract and shared across all participants, we did not include random slopes. Instead, to account for variation in baseline comprehension across participants and differences in abstract difficult, we included random intercepts for both participants and abstracts. Specifically, we specified random intercepts for \texttt{participant\_id} using the \texttt{groups} argument in \texttt{statsmodels}, and modeled independent variance components for \texttt{abstract\_id} using the \texttt{vc\_formula}. This modeling approach controls for both individual-level and content-level heterogeneity while estimating the overall association between each metric and comprehension outcomes. For comparability, all human ratings and automated scores were standardized to have zero mean and unit variance prior to model fitting. $p$-values were adjusted for multiple comparisons using the Benjamini--Hochberg procedure 
(\texttt{fdr\_bh} in \texttt{statsmodels.stats.multitest.multipletests}). Metrics with $p$-values less than 0.05 are considered statistically significant.

\makeatletter
\renewcommand{\@biblabel}[1]{\hfill #1.}
\makeatother

\bibliography{amia}  

\begin{thebibliography}{10}
\urlstyle{rm}
\expandafter\ifx\csname url\endcsname\relax
  \def\url#1{\texttt{#1}}\fi
\expandafter\ifx\csname urlprefix\endcsname\relax\def\urlprefix{URL }\fi
\expandafter\ifx\csname doiprefix\endcsname\relax\def\doiprefix{DOI: }\fi
\providecommand{\bibinfo}[2]{#2}
\providecommand{\eprint}[2][]{\url{#2}}

\bibitem{ravinetto2023responsible}
\bibinfo{author}{Ravinetto, R.} \& \bibinfo{author}{Singh, J.~A.}
\newblock \bibinfo{journal}{\bibinfo{title}{Responsible dissemination of health and medical research: some guidance points}}.
\newblock {\emph{\JournalTitle{BMJ evidence-based medicine}}} \textbf{\bibinfo{volume}{28}}, \bibinfo{pages}{144--147} (\bibinfo{year}{2023}).

\bibitem{NIH2025}
\bibinfo{author}{{National Institutes of Health}}.
\newblock \bibinfo{title}{Clearly communicating research results across the clinical trials continuum} (\bibinfo{year}{2025}).
\newblock \bibinfo{note}{Accessed: March 10, 2025}.

\bibitem{Guo2022RetrievalAO}
\bibinfo{author}{Guo, Y.}, \bibinfo{author}{Qiu, W.}, \bibinfo{author}{Leroy, G.}, \bibinfo{author}{Wang, S.} \& \bibinfo{author}{Cohen, T.~A.}
\newblock \bibinfo{journal}{\bibinfo{title}{Retrieval augmentation of large language models for lay language generation}}.
\newblock {\emph{\JournalTitle{Journal of biomedical informatics}}} \bibinfo{pages}{104580} (\bibinfo{year}{2022}).

\bibitem{luhn1958automatic}
\bibinfo{author}{Luhn, H.~P.}
\newblock \bibinfo{journal}{\bibinfo{title}{The automatic creation of literature abstracts}}.
\newblock {\emph{\JournalTitle{IBM Journal of research and development}}} \textbf{\bibinfo{volume}{2}}, \bibinfo{pages}{159--165} (\bibinfo{year}{1958}).

\bibitem{lewis2019bart}
\bibinfo{author}{Lewis, M.} \emph{et~al.}
\newblock \bibinfo{journal}{\bibinfo{title}{Bart: Denoising sequence-to-sequence pre-training for natural language generation, translation, and comprehension}}.
\newblock {\emph{\JournalTitle{arXiv preprint arXiv:1910.13461}}}  (\bibinfo{year}{2019}).

\bibitem{guo2021automated}
\bibinfo{author}{Guo, Y.}, \bibinfo{author}{Qiu, W.}, \bibinfo{author}{Wang, Y.} \& \bibinfo{author}{Cohen, T.}
\newblock \bibinfo{title}{Automated lay language summarization of biomedical scientific reviews}.
\newblock In \emph{\bibinfo{booktitle}{Proceedings of the AAAI Conference on Artificial Intelligence}}, vol.~\bibinfo{volume}{35}, \bibinfo{pages}{160--168} (\bibinfo{year}{2021}).

\bibitem{devaraj2021paragraph}
\bibinfo{author}{Devaraj, A.}, \bibinfo{author}{Marshall, I.}, \bibinfo{author}{Wallace, B.~C.} \& \bibinfo{author}{Li, J.~J.}
\newblock \bibinfo{title}{Paragraph-level simplification of medical texts}.
\newblock In \emph{\bibinfo{booktitle}{Proceedings of the 2021 Conference of the North American Chapter of the Association for Computational Linguistics: Human Language Technologies}}, \bibinfo{pages}{4972--4984} (\bibinfo{year}{2021}).

\bibitem{touvron2023llama}
\bibinfo{author}{Touvron, H.} \emph{et~al.}
\newblock \bibinfo{journal}{\bibinfo{title}{Llama 2: Open foundation and fine-tuned chat models}}.
\newblock {\emph{\JournalTitle{arXiv preprint arXiv:2307.09288}}}  (\bibinfo{year}{2023}).

\bibitem{luo2023chatgpt}
\bibinfo{author}{Luo, Z.}, \bibinfo{author}{Xie, Q.} \& \bibinfo{author}{Ananiadou, S.}
\newblock \bibinfo{journal}{\bibinfo{title}{Chatgpt as a factual inconsistency evaluator for abstractive text summarization}}.
\newblock {\emph{\JournalTitle{arXiv preprint arXiv:2303.15621}}}  (\bibinfo{year}{2023}).

\bibitem{you2024uiuc_bionlp}
\bibinfo{author}{You, Z.}, \bibinfo{author}{Radhakrishna, S.}, \bibinfo{author}{Ming, S.} \& \bibinfo{author}{Kilicoglu, H.}
\newblock \bibinfo{title}{Uiuc\_bionlp at biolaysumm: an extract-then-summarize approach augmented with wikipedia knowledge for biomedical lay summarization}.
\newblock In \emph{\bibinfo{booktitle}{Proceedings of the 23rd Workshop on Biomedical Natural Language Processing}}, \bibinfo{pages}{132--143} (\bibinfo{year}{2024}).

\bibitem{goldsack2024overview}
\bibinfo{author}{Goldsack, T.}, \bibinfo{author}{Scarton, C.}, \bibinfo{author}{Shardlow, M.} \& \bibinfo{author}{Lin, C.}
\newblock \bibinfo{journal}{\bibinfo{title}{Overview of the biolaysumm 2024 shared task on the lay summarization of biomedical research articles}}.
\newblock {\emph{\JournalTitle{arXiv preprint arXiv:2408.08566}}}  (\bibinfo{year}{2024}).

\bibitem{leroy2024text}
\bibinfo{author}{Leroy, G.}, \bibinfo{author}{Kauchak, D.}, \bibinfo{author}{Harber, P.}, \bibinfo{author}{Pal, A.} \& \bibinfo{author}{Shukla, A.}
\newblock \bibinfo{journal}{\bibinfo{title}{Text and audio simplification: Human vs. chatgpt}}.
\newblock {\emph{\JournalTitle{AMIA Summits on Translational Science Proceedings}}} \textbf{\bibinfo{volume}{2024}}, \bibinfo{pages}{295} (\bibinfo{year}{2024}).

\bibitem{cohen2025coherence}
\bibinfo{author}{Cohen, T.}, \bibinfo{author}{Xu, W.}, \bibinfo{author}{Guo, Y.}, \bibinfo{author}{Pakhomov, S.} \& \bibinfo{author}{Leroy, G.}
\newblock \bibinfo{journal}{\bibinfo{title}{Coherence and comprehensibility: Large language models predict lay understanding of health-related content}}.
\newblock {\emph{\JournalTitle{Journal of Biomedical Informatics}}} \textbf{\bibinfo{volume}{161}}, \bibinfo{pages}{104758} (\bibinfo{year}{2025}).

\bibitem{jain2021summarization}
\bibinfo{author}{Jain, D.}, \bibinfo{author}{Borah, M.~D.} \& \bibinfo{author}{Biswas, A.}
\newblock \bibinfo{journal}{\bibinfo{title}{Summarization of legal documents: Where are we now and the way forward}}.
\newblock {\emph{\JournalTitle{Computer Science Review}}} \textbf{\bibinfo{volume}{40}}, \bibinfo{pages}{100388} (\bibinfo{year}{2021}).

\bibitem{ondov2022survey}
\bibinfo{author}{Ondov, B.}, \bibinfo{author}{Attal, K.} \& \bibinfo{author}{Demner-Fushman, D.}
\newblock \bibinfo{journal}{\bibinfo{title}{A survey of automated methods for biomedical text simplification}}.
\newblock {\emph{\JournalTitle{Journal of the American Medical Informatics Association}}} \textbf{\bibinfo{volume}{29}}, \bibinfo{pages}{1976--1988} (\bibinfo{year}{2022}).

\bibitem{lin2004rouge}
\bibinfo{author}{Lin, C.-Y.}
\newblock \bibinfo{title}{Rouge: A package for automatic evaluation of summaries}.
\newblock In \emph{\bibinfo{booktitle}{Text summarization branches out}}, \bibinfo{pages}{74--81} (\bibinfo{year}{2004}).

\bibitem{sulem2018bleu}
\bibinfo{author}{Sulem, E.}, \bibinfo{author}{Abend, O.} \& \bibinfo{author}{Rappoport, A.}
\newblock \bibinfo{journal}{\bibinfo{title}{Bleu is not suitable for the evaluation of text simplification}}.
\newblock {\emph{\JournalTitle{arXiv preprint arXiv:1810.05995}}}  (\bibinfo{year}{2018}).

\bibitem{Guo2023APPLSEE}
\bibinfo{author}{Guo, Y.}, \bibinfo{author}{August, T.}, \bibinfo{author}{Leroy, G.}, \bibinfo{author}{Cohen, T.~A.} \& \bibinfo{author}{Wang, L.~L.}
\newblock \bibinfo{title}{Appls: Evaluating evaluation metrics for plain language summarization}.
\newblock In \emph{\bibinfo{booktitle}{Conference on Empirical Methods in Natural Language Processing}} (\bibinfo{year}{2023}).

\bibitem{hirschberg2003experiments}
\bibinfo{author}{Hirschberg, J.}, \bibinfo{author}{Liscombe, J.} \& \bibinfo{author}{Venditti, J.}
\newblock \bibinfo{title}{Experiments in emotional speech}.
\newblock In \emph{\bibinfo{booktitle}{ISCA \& IEEE Workshop on Spontaneous Speech Processing and Recognition}}, \bibinfo{pages}{1--7} (\bibinfo{year}{2003}).

\bibitem{wiebe2004learning}
\bibinfo{author}{Wiebe, J.}, \bibinfo{author}{Wilson, T.}, \bibinfo{author}{Bruce, R.}, \bibinfo{author}{Bell, M.} \& \bibinfo{author}{Martin, M.}
\newblock \bibinfo{journal}{\bibinfo{title}{Learning subjective language}}.
\newblock {\emph{\JournalTitle{Computational linguistics}}} \textbf{\bibinfo{volume}{30}}, \bibinfo{pages}{277--308} (\bibinfo{year}{2004}).

\bibitem{haselton2015evolution}
\bibinfo{author}{Haselton, M.~G.}, \bibinfo{author}{Nettle, D.} \& \bibinfo{author}{Andrews, P.~W.}
\newblock \bibinfo{journal}{\bibinfo{title}{The evolution of cognitive bias}}.
\newblock {\emph{\JournalTitle{The handbook of evolutionary psychology}}} \bibinfo{pages}{724--746} (\bibinfo{year}{2015}).

\bibitem{guo2022cells}
\bibinfo{author}{Guo, Y.}, \bibinfo{author}{Qiu, W.}, \bibinfo{author}{Leroy, G.}, \bibinfo{author}{Wang, S.} \& \bibinfo{author}{Cohen, T.}
\newblock \bibinfo{journal}{\bibinfo{title}{Cells: A parallel corpus for biomedical lay language generation}}.
\newblock {\emph{\JournalTitle{arXiv preprint arXiv:2211.03818}}}  (\bibinfo{year}{2022}).

\bibitem{beck2023quality}
\bibinfo{author}{Beck, J.}
\newblock \bibinfo{journal}{\bibinfo{title}{Quality aspects of annotated data: A research synthesis}}.
\newblock {\emph{\JournalTitle{AStA Wirtschafts-und Sozialstatistisches Archiv}}} \textbf{\bibinfo{volume}{17}}, \bibinfo{pages}{331--353} (\bibinfo{year}{2023}).

\bibitem{leroy2018next}
\bibinfo{author}{Leroy, G.} \emph{et~al.}
\newblock \bibinfo{journal}{\bibinfo{title}{Next-generation metrics for monitoring genetic erosion within populations of conservation concern}}.
\newblock {\emph{\JournalTitle{Evolutionary Applications}}} \textbf{\bibinfo{volume}{11}}, \bibinfo{pages}{1066--1083} (\bibinfo{year}{2018}).

\bibitem{srikanth2020elaborative}
\bibinfo{author}{Srikanth, N.} \& \bibinfo{author}{Li, J.~J.}
\newblock \bibinfo{journal}{\bibinfo{title}{Elaborative simplification: Content addition and explanation generation in text simplification}}.
\newblock {\emph{\JournalTitle{arXiv preprint arXiv:2010.10035}}}  (\bibinfo{year}{2020}).

\bibitem{matz2024potential}
\bibinfo{author}{Matz, S.~C.} \emph{et~al.}
\newblock \bibinfo{journal}{\bibinfo{title}{The potential of generative ai for personalized persuasion at scale}}.
\newblock {\emph{\JournalTitle{Scientific Reports}}} \textbf{\bibinfo{volume}{14}}, \bibinfo{pages}{4692} (\bibinfo{year}{2024}).

\bibitem{karinshak2023working}
\bibinfo{author}{Karinshak, E.}, \bibinfo{author}{Liu, S.~X.}, \bibinfo{author}{Park, J.~S.} \& \bibinfo{author}{Hancock, J.~T.}
\newblock \bibinfo{journal}{\bibinfo{title}{Working with ai to persuade: Examining a large language model's ability to generate pro-vaccination messages}}.
\newblock {\emph{\JournalTitle{Proceedings of the ACM on Human-Computer Interaction}}} \textbf{\bibinfo{volume}{7}}, \bibinfo{pages}{1--29} (\bibinfo{year}{2023}).

\bibitem{rakedzon2017automatic}
\bibinfo{author}{Rakedzon, T.}, \bibinfo{author}{Segev, E.}, \bibinfo{author}{Chapnik, N.}, \bibinfo{author}{Yosef, R.} \& \bibinfo{author}{Baram-Tsabari, A.}
\newblock \bibinfo{journal}{\bibinfo{title}{Automatic jargon identifier for scientists engaging with the public and science communication educators}}.
\newblock {\emph{\JournalTitle{PloS one}}} \textbf{\bibinfo{volume}{12}}, \bibinfo{pages}{e0181742} (\bibinfo{year}{2017}).

\bibitem{martinez2020can}
\bibinfo{author}{Mart{\'\i}nez~Silvagnoli, L.}, \bibinfo{author}{Shepherd, C.}, \bibinfo{author}{Pritchett, J.} \& \bibinfo{author}{Gardner, J.}
\newblock \bibinfo{journal}{\bibinfo{title}{How can we optimize the readability and format of plain language summaries for medical journal articles? a cross-sectional survey study}}.
\newblock {\emph{\JournalTitle{J Med Internet Res}}} \textbf{\bibinfo{volume}{22122}} (\bibinfo{year}{2020}).

\bibitem{brehaut2011presenting}
\bibinfo{author}{Brehaut, J.~C.} \emph{et~al.}
\newblock \bibinfo{journal}{\bibinfo{title}{Presenting evidence to patients online: what do web users think of consumer summaries of cochrane musculoskeletal reviews?}}
\newblock {\emph{\JournalTitle{Journal of Medical Internet Research}}} \textbf{\bibinfo{volume}{13}}, \bibinfo{pages}{e1532} (\bibinfo{year}{2011}).

\bibitem{buljan2020comparison}
\bibinfo{author}{Buljan, I.} \emph{et~al.}
\newblock \bibinfo{journal}{\bibinfo{title}{Comparison of blogshots with plain language summaries of cochrane systematic reviews: a qualitative study and randomized trial}}.
\newblock {\emph{\JournalTitle{Trials}}} \textbf{\bibinfo{volume}{21}}, \bibinfo{pages}{1--10} (\bibinfo{year}{2020}).

\bibitem{buljan2020framing}
\bibinfo{author}{Buljan, I.} \emph{et~al.}
\newblock \bibinfo{journal}{\bibinfo{title}{Framing the numerical findings of cochrane plain language summaries: two randomized controlled trials}}.
\newblock {\emph{\JournalTitle{BMC medical research methodology}}} \textbf{\bibinfo{volume}{20}}, \bibinfo{pages}{1--9} (\bibinfo{year}{2020}).

\bibitem{santesso2015summary}
\bibinfo{author}{Santesso, N.} \emph{et~al.}
\newblock \bibinfo{journal}{\bibinfo{title}{A summary to communicate evidence from systematic reviews to the public improved understanding and accessibility of information: a randomized controlled trial}}.
\newblock {\emph{\JournalTitle{Journal of Clinical Epidemiology}}} \textbf{\bibinfo{volume}{68}}, \bibinfo{pages}{182--190}, \doiprefix\url{10.1016/j.jclinepi.2014.04.009} (\bibinfo{year}{2015}).
\newblock \bibinfo{note}{Epub 2014 Jul 14}.

\bibitem{bredbenner2019video}
\bibinfo{author}{Bredbenner, K.} \& \bibinfo{author}{Simon, S.~M.}
\newblock \bibinfo{journal}{\bibinfo{title}{Video abstracts and plain language summaries are more effective than graphical abstracts and published abstracts}}.
\newblock {\emph{\JournalTitle{PloS one}}} \textbf{\bibinfo{volume}{14}}, \bibinfo{pages}{e0224697} (\bibinfo{year}{2019}).

\bibitem{buljan2018no}
\bibinfo{author}{Buljan, I.} \emph{et~al.}
\newblock \bibinfo{journal}{\bibinfo{title}{No difference in knowledge obtained from infographic or plain language summary of a cochrane systematic review: three randomized controlled trials}}.
\newblock {\emph{\JournalTitle{Journal of clinical epidemiology}}} \textbf{\bibinfo{volume}{97}}, \bibinfo{pages}{86--94} (\bibinfo{year}{2018}).

\bibitem{kirkpatrick2017understanding}
\bibinfo{author}{Kirkpatrick, E.} \emph{et~al.}
\newblock \bibinfo{journal}{\bibinfo{title}{Understanding plain english summaries. a comparison of two approaches to improve the quality of plain english summaries in research reports}}.
\newblock {\emph{\JournalTitle{Research involvement and engagement}}} \textbf{\bibinfo{volume}{3}}, \bibinfo{pages}{1--14} (\bibinfo{year}{2017}).

\bibitem{OpenAI2024GPT4}
\bibinfo{author}{OpenAI}.
\newblock \bibinfo{title}{Gpt-4 technical report}.
\newblock \bibinfo{howpublished}{\url{https://openai.com/research/gpt-4}} (\bibinfo{year}{2024}).
\newblock \bibinfo{note}{Accessed: February 12, 2024}.

\bibitem{panickssery2024llm}
\bibinfo{author}{Panickssery, A.}, \bibinfo{author}{Bowman, S.} \& \bibinfo{author}{Feng, S.}
\newblock \bibinfo{journal}{\bibinfo{title}{Llm evaluators recognize and favor their own generations}}.
\newblock {\emph{\JournalTitle{Advances in Neural Information Processing Systems}}} \textbf{\bibinfo{volume}{37}}, \bibinfo{pages}{68772--68802} (\bibinfo{year}{2024}).

\bibitem{papineni2002bleu}
\bibinfo{author}{Papineni, K.}, \bibinfo{author}{Roukos, S.}, \bibinfo{author}{Ward, T.} \& \bibinfo{author}{Zhu, W.-J.}
\newblock \bibinfo{title}{Bleu: a method for automatic evaluation of machine translation}.
\newblock In \emph{\bibinfo{booktitle}{Proceedings of the 40th annual meeting of the Association for Computational Linguistics}}, \bibinfo{pages}{311--318} (\bibinfo{year}{2002}).

\bibitem{banerjee2005meteor}
\bibinfo{author}{Banerjee, S.} \& \bibinfo{author}{Lavie, A.}
\newblock \bibinfo{title}{Meteor: An automatic metric for mt evaluation with improved correlation with human judgments}.
\newblock In \emph{\bibinfo{booktitle}{Proceedings of the acl workshop on intrinsic and extrinsic evaluation measures for machine translation and/or summarization}}, \bibinfo{pages}{65--72} (\bibinfo{year}{2005}).

\bibitem{xu2016optimizing}
\bibinfo{author}{Xu, W.}, \bibinfo{author}{Napoles, C.}, \bibinfo{author}{Pavlick, E.}, \bibinfo{author}{Chen, Q.} \& \bibinfo{author}{Callison-Burch, C.}
\newblock \bibinfo{journal}{\bibinfo{title}{Optimizing statistical machine translation for text simplification}}.
\newblock {\emph{\JournalTitle{Transactions of the Association for Computational Linguistics}}} \textbf{\bibinfo{volume}{4}}, \bibinfo{pages}{401--415} (\bibinfo{year}{2016}).

\bibitem{zhang2019bertscore}
\bibinfo{author}{Zhang, T.}, \bibinfo{author}{Kishore, V.}, \bibinfo{author}{Wu, F.}, \bibinfo{author}{Weinberger, K.~Q.} \& \bibinfo{author}{Artzi, Y.}
\newblock \bibinfo{journal}{\bibinfo{title}{Bertscore: Evaluating text generation with bert}}.
\newblock {\emph{\JournalTitle{arXiv preprint arXiv:1904.09675}}}  (\bibinfo{year}{2019}).

\bibitem{maddela2022lens}
\bibinfo{author}{Maddela, M.}, \bibinfo{author}{Dou, Y.}, \bibinfo{author}{Heineman, D.} \& \bibinfo{author}{Xu, W.}
\newblock \bibinfo{journal}{\bibinfo{title}{Lens: A learnable evaluation metric for text simplification}}.
\newblock {\emph{\JournalTitle{arXiv preprint arXiv:2212.09739}}}  (\bibinfo{year}{2022}).

\bibitem{deutsch2021towards}
\bibinfo{author}{Deutsch, D.}, \bibinfo{author}{Bedrax-Weiss, T.} \& \bibinfo{author}{Roth, D.}
\newblock \bibinfo{journal}{\bibinfo{title}{Towards question-answering as an automatic metric for evaluating the content quality of a summary}}.
\newblock {\emph{\JournalTitle{Transactions of the Association for Computational Linguistics}}} \textbf{\bibinfo{volume}{9}}, \bibinfo{pages}{774--789} (\bibinfo{year}{2021}).

\bibitem{fabbri2021summeval}
\bibinfo{author}{Fabbri, A.~R.} \emph{et~al.}
\newblock \bibinfo{journal}{\bibinfo{title}{Summeval: Re-evaluating summarization evaluation}}.
\newblock {\emph{\JournalTitle{Transactions of the Association for Computational Linguistics}}} \textbf{\bibinfo{volume}{9}}, \bibinfo{pages}{391--409} (\bibinfo{year}{2021}).

\bibitem{alva-manchego-etal-2019-easse}
\bibinfo{author}{Alva-Manchego, F.}, \bibinfo{author}{Martin, L.}, \bibinfo{author}{Scarton, C.} \& \bibinfo{author}{Specia, L.}
\newblock \bibinfo{title}{{EASSE}: {E}asier automatic sentence simplification evaluation}.
\newblock In \emph{\bibinfo{booktitle}{Proceedings of the 2019 Conference on Empirical Methods in Natural Language Processing and the 9th International Joint Conference on Natural Language Processing (EMNLP-IJCNLP): System Demonstrations}}, \bibinfo{pages}{49--54}, \doiprefix\url{10.18653/v1/D19-3009} (\bibinfo{publisher}{Association for Computational Linguistics}, \bibinfo{address}{Hong Kong, China}, \bibinfo{year}{2019}).

\end{thebibliography}



\section*{Appendix}

\begin{table}[ht]
\centering
\caption{Mixed-effects model results for human ratings predicting multiple-choice question comprehension scores (N = 1346). $p$-values are adjusted for multiple comparisons using the Benjamini--Hochberg correction. Metrics with $p$-values less than 0.05 are considered statistically significant.}
\begin{tabular}{lccccc}
\toprule
\textbf{Metric} & \textbf{Coef.} & \textbf{$p$-value} & \textbf{95\% CI Lower} & \textbf{95\% CI Upper} & \textbf{Significant} \\
\midrule
Faithfulness     & 0.040 & 0.000 & 0.022 & 0.058 & Yes \\
Simplicity       & 0.034 & 0.001 & 0.014 & 0.053 & Yes \\
Background       & 0.033 & 0.001 & 0.014 & 0.052 & Yes \\
Coherence        & 0.033 & 0.001 & 0.014 & 0.051 & Yes \\
Informativeness  & 0.026 & 0.006 & 0.008 & 0.045 & Yes \\
\bottomrule
\end{tabular}
\label{rating_mcq_mixed_whole_bh}
\end{table}

\begin{table}[ht]
\centering
\caption{Mixed-effects model results for automated metrics predicting normalized MCQ comprehension scores (N = 1346). $p$-values are adjusted for multiple comparisons using the Benjamini--Hochberg correction. Metrics with $p$-values less than 0.05 are considered statistically significant.}
\begin{tabular}{lccccc}
\toprule
\textbf{Metric} & \textbf{Coef.} & \textbf{$p$-value} & \textbf{95\% CI Lower} & \textbf{95\% CI Upper} & \textbf{Significant} \\
\midrule
QAEval         & 0.040 & 0.000 & 0.026 & 0.055 & Yes \\
Coherence-chn  & 0.034 & 0.000 & 0.019 & 0.049 & Yes \\
BERTScore      & 0.028 & 0.001 & 0.013 & 0.043 & Yes \\
Coherence-bag  & 0.024 & 0.004 & 0.009 & 0.039 & Yes \\
ROUGE          & 0.017 & 0.002 & 0.002 & 0.032 & Yes \\
METEOR         & 0.016 & 0.057 & 0.001 & 0.031 & Yes \\
BLEU           & 0.009 & 0.367 & -0.006 & 0.024 & No \\
LENS           & 0.008 & 0.403 & -0.007 & 0.023 & No \\
GPT-PPL        & 0.001 & 0.849 & -0.014 & 0.017 & No \\
SARI           & -0.003 & 0.761 & -0.018 & 0.012 & No \\
\bottomrule
\end{tabular}
\label{auto_mcq_mixed_whole_bh}
\end{table}

\end{document}